\documentclass[10pt,twocolumn,letterpaper]{article}

\usepackage{wacv}
\usepackage{times}
\usepackage{epsfig}
\usepackage{graphicx}
\usepackage{amsmath}
\usepackage{amssymb}
\usepackage{booktabs}
\usepackage{algorithm}
\usepackage{algorithmic}
\usepackage{booktabs}
\usepackage{soul}
\usepackage{arydshln}
\usepackage{bbm}
\usepackage{subcaption}
\usepackage{wrapfig}
\usepackage{makecell}
\usepackage{xurl}
\usepackage[accsupp]{axessibility}
\usepackage{array}  

\def\wacvPaperID{1344} 

\wacvalgorithmstrack   

\wacvfinalcopy 

\ifwacvfinal
\usepackage[breaklinks=true,bookmarks=false]{hyperref}
\else
\usepackage[pagebackref=true,breaklinks=true,colorlinks,bookmarks=false]{hyperref}
\fi

\begin{document}

\title{Exploiting Long-Term Dependencies for Generating Dynamic Scene Graphs}



\author{Shengyu Feng$^{1}$\thanks{Work partially done during an internship at Intel Labs.} 
\quad Hesham Mostafa$^2$ 
\quad Marcel Nassar$^2$ 
\quad Somdeb Majumdar$^2$
\quad Subarna Tripathi$^2$ 
\\
$^1$Carnegie Mellon University
\quad\quad $^2$Intel Labs\\
{\tt\small shengyuf@andrew.cmu.edu}, \\\tt{\small\{hesham.mostafa, marcel.nassar, somdeb.majumdar, subarna.tripathi\}@intel.com}
}
\maketitle

\thispagestyle{empty}

\begin{abstract}
Dynamic scene graph generation from a video is challenging due to the temporal dynamics of the scene and the inherent temporal fluctuations of predictions. We hypothesize that capturing long-term temporal dependencies is the key to effective generation of dynamic scene graphs. 
We propose to learn the long-term dependencies in a video by capturing the object-level consistency and inter-object relationship dynamics over object-level long-term tracklets using transformers. 
Experimental results demonstrate that our \emph{Dynamic Scene Graph Detection Transformer} (DSG-DETR) outperforms state-of-the-art methods by a significant margin
on the benchmark dataset Action Genome. Our ablation studies validate the effectiveness of each component of the proposed approach. 
The source code is available at \url{https://github.com/Shengyu-Feng/DSG-DETR}.


\end{abstract}

\section{Introduction}

A scene graph is a directed graph where each node represents a labelled object 
and each edge represents an inter-object relationship, also known as a \textit{predicate}.
Learning visual relations in static images is a difficult problem due to its combinatorial nature. 
The underlying spatio-temporal dynamics and temporal fluctuations of predictions
make the dynamic scene graph generation from video even harder.  
The naive solution to dynamic scene graph generation is simply applying the static scene graph generation method on each video frame without considering the temporal context. Recently a line of work~\cite{UnifiedGraphStructured2021,STTran_2021,TargetAdaptiveContext_2021,Li_2022_CVPR} emerged that demonstrated the importance of capturing the spatial as well as the temporal dependencies for dynamic scene graph generations.    

\begin{figure}
    \centering
    \includegraphics[width=.8\linewidth]{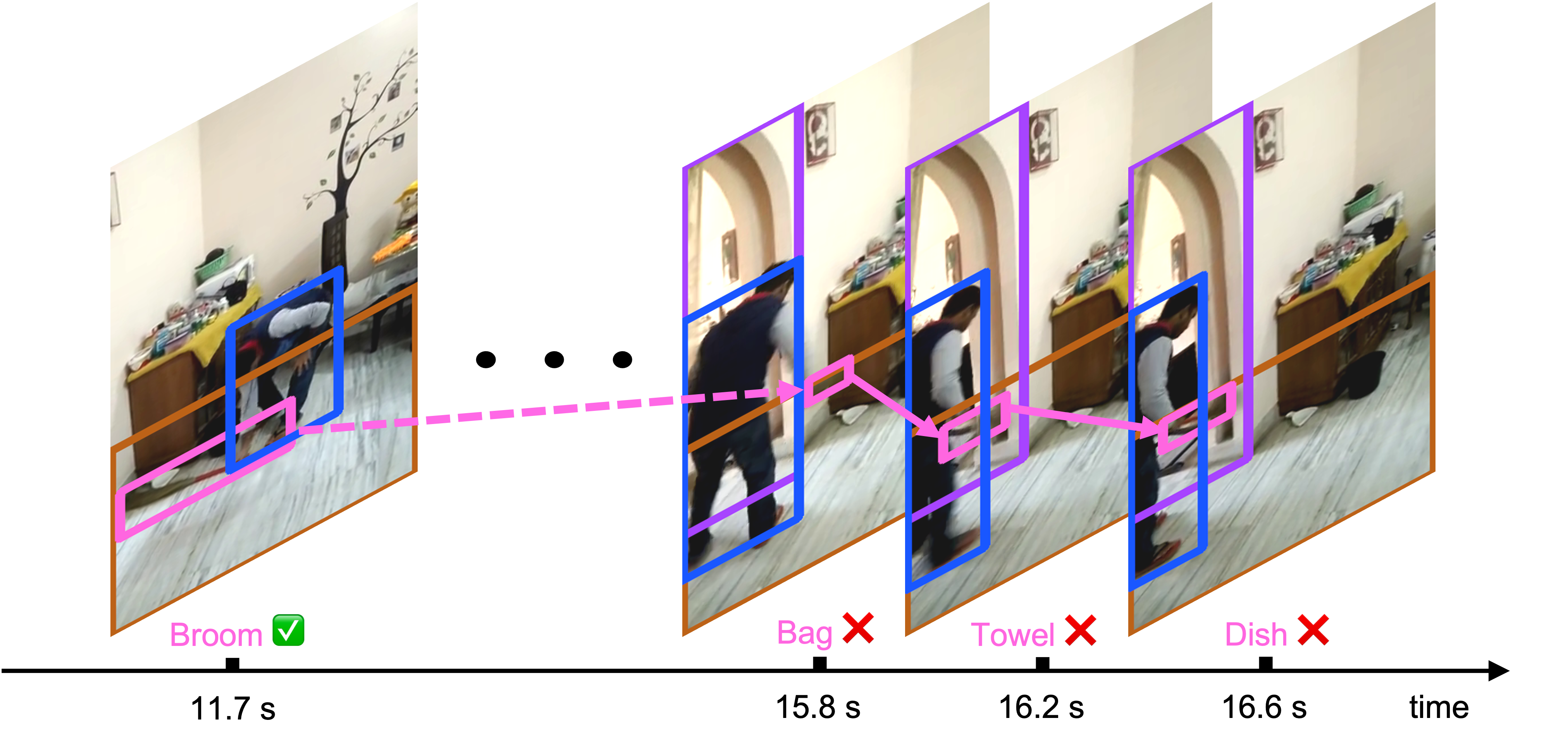}
    \caption{An example where the short-term temporal dependencies fail.
    The broom, bounded by the pink bounding boxes, is quite challenging to recognize on the rightmost three frames. Previous methods capturing only the short-term dependencies (indicated by the solid pink arrows) fail to make the correct prediction, while our method capturing the long-term dependencies (over more than 4 seconds), can recognize the broom and predict the human-broom relationship. 
    }
    \label{fig:example}
    \vspace{-10pt}
\end{figure}

The predominant ways to realize spatio-temporal consistencies focus on the construction of the spatio-temporal graph. Arnab \textit{et al.} \cite{UnifiedGraphStructured2021} construct a unified graph structure, utilizing a fully connected graph over the foreground nodes from the frames in a sliding window, and the connections between the foreground nodes and context nodes in each frame. Although this spatio-temporal graph can successfully perform message passing over both the spatial and temporal domains, such a fully connected graph is computationally expensive. 
In practice, such models resort to using a small sliding window consisting of 3 to 5 key frames, making them incapable of reasoning over long-term sequences. 
Another recent work, Spatial-Temporal Transformer (STTran)  \cite{STTran_2021} grounds the model on the adjacent key frames. As a result, these methods can only achieve short-term consistency and fail to capture long-term dependencies. 

Fig \ref{fig:example} shows an example where occlusion and fast movement make it extremely difficult for any static image-based object detector to recognize the broom (bounded by the pink bounding box) in the rightmost three video frames. 
Any model that relies on capturing only short-term dependencies will fail to detect objects correctly in scenarios such as this example. This will result in incorrect dynamic scene graph generation. Predictions in frames, where an object might be occluded, may be improved by leveraging correct predictions from frames where it is easily detectable and recognizable. Since such ``good" frames may be several frames away in the past or the future (4 seconds in this example), capturing long-term dependencies is crucial to improving the overall scene graph generation performance.

In this paper, we study the benefits of utilizing long-term temporal dependencies for objects and relations in dynamic scene graph generation tasks.
We quantify it via estimating the hypothetical best case by exploiting the ground-truth object-tracks information when evaluated on scene graph generation tasks using Recall metric. This hypothetical best case significantly outperforms existing methods on dynamic scene graph generation. 
Next, we propose a paradigm 
for consistent video object detections without the access to ground truth object tracks. To this end, we construct the temporal sequences by tracking each object instance using the Hungarian matching algorithm \cite{Kuhn55thehungarian},  
and apply a transformer encoder to leverage the temporal consistency within all such sequences. We also model the relationship transitions through the sequences of predicated subject-object classes using another transformer network. 
Our framework, named \textbf{DSG-DETR} (Dynamic Scene Graph Detection Transformer), 
performs comparably with the hypothetical best case explained above. 
The experimental results on the Action Genome dataset ~\cite{Action_genome_Ji_2020_CVPR} also demonstrate that \textbf{DSG-DETR} can achieve significant improvements over the state-of-the-art methods for video scene graph generation. 

\noindent
The main contributions of our work can be summarized as:
\begin{itemize}
    \item We hypothesize that the key to improving dynamic scene graph generation is in capturing \emph{long-term} temporal dependencies of objects and visual relationships. 
    \item We quantify the benefit of capturing long-term dependencies by estimating the hypothetical best case (an upper-bound) on the scene graph generation task performances by utilizing the ground-truth object tracks. 
    \item We then show that by 
    capturing object consistencies and inter-object relationship dynamics within predicted object tracklets,
    our method DSG-DETR approaches the hypothetical best case performance. 
    \item DSG-DETR significantly outperforms existing state-of-the-art methods on the video scene graph generation benchmark dataset Action Genome.  
\end{itemize}

\section{Related work}
\paragraph{Scene graph generation.}
Scene graph generation (SGG) has become an important problem in computer vision since Johnson \textit{et al.}~\cite{image_retrieval_SG_2015} introduced the concept of graph-based image representation. 
A large body of work has focused on scene graph generation from  images~\cite{neural_motifs_2017,Li2017SceneGG,VCTree_Tang_2019_CVPR,zhang2019vrd,lu2016visual,lin2020gps}. These methods focus on either sophisticated architecture design or contextual feature fusion strategies, such as message passing or recurrent neural networks, to optimize SGG performance on the image scene graph benchmark dataset~\cite{krishnavisualgenome}.
Such static SGG methods do not consider the dynamics of a video. 
Video SGG is significantly more challenging than image SGG due to the underlying spatio-temporal dynamics involving objects and inter-object relationships.   
A line of recent and concurrent work ~\cite{UnifiedGraphStructured2021,STTran_2021,TargetAdaptiveContext_2021,HOI_video_2021} looks at the problem of video scene graph generation via modeling the spatio-temporal dynamics of relationships. 
While ~\cite{UnifiedGraphStructured2021} takes an approach of message passing in a  spatio-temporal graph for capturing the relationship dynamics, others rely on visual transformers for it. Some recent works \cite{Jung2021,Gao2021,TRACE} also utilize the tracking to boost the temporal context aggregation.  \cite{Jung2021} and \cite{Gao2021} are based on the track-to-detect paradigm which first tracks the objects across the whole video and then figures out the pairwise relationship among tracklet pairs. However, this paradigm is highly sensitive to the tracking results and not flexible for the frame-level scene graph generation. TRACE \cite{TRACE}, in contrast, utilizes a detect-to-track paradigm, but it is still limited to the short-term dependency and faces the aggregation problem of different prediction results for the same frame in different video segments.  One concurrent work~\cite{Li_2022_CVPR} leverages anticipatory prediction as pre-training and combines it with the fine-tuning strategies. 
Our coarse tracking method assimilates the complementary strengths of tracking based methods to allow the flexibility of long-term dependencies and frame-level prediction. \vspace{-5mm}

\paragraph{Transformer models in video analysis.}
Following the immense success of transformers ~\cite{vaswani2017attention} in natural language processing, they have been shown to be effective for image perception tasks~\cite{dosovitskiy2020vit,DETRCarionMSUKZ20,pmlr_v80_parmar18a} and video understanding tasks~\cite{videoBERT_Sun_2019_ICCV,GirdharCDZ19,eccv_GarciaN20}. 
A recent work studies transformers for video SGG~\cite{STTran_2021}, which is also the theme of this paper. Another closely related problem is Human-Object-Interaction (HOI) detection from video where a Human-Object Relationship transformer has been utilized~\cite{HOI_video_2021}. 

We hypothesize and experimentally show that the temporal fluctuation of object-level predictions hinders the performance of dynamic SGG tasks significantly. While~\cite{STTran_2021} aims to capture the relationship dynamics via spatial encoder and temporal decoder, the impact of temporally consistent object predictions largely remain unaddressed. 
To the best of our knowledge, none of the methods aims to systematically capture the long-term dynamics at object-level.
To summarize, we utilize a long-term temporal dependency via an online tracklet construction framework.
Next, an object-centric transformer is employed on these sequences resulting in temporally consistent object recognition, followed by a spatio-temporal relationship transformer on the predicted sequences of the same subject-object classes. 



\section{Problem statement and notations}

\subsection{Dynamic scene graph generation}
Given a video as a sequence of $I$ key frames, 
we want to predict the objects for each frame, in terms of their positions and classes, and the relationships among them.
Use $\mathcal{C}$ and $\mathcal{P}$ to denote the object class set and the predicate set respectively.
We define each object as a tuple comprising its bounding box $\mathbf{b}$ and object class $\mathbf{c}$, i.e., $\mathcal{O}=\langle\mathbf{b}, \mathbf{c}\rangle$. Here, $\mathbf{b}\in[0,1]^4$ is a vector composed of the object center coordinates and its width and height relative to the image size.
 $\mathbf{c}\in\{0,1\}^{|\mathcal{C}|}$ is a one-hot vector with $\mathbf{c}[i]=1$ and all other dimensions $0$, where the $i$-th element of $\mathcal{C}$ corresponds to the class of this object.
 
The relationship tuple for a subject-object pair is defined as $\langle \mathcal{O}_s, p, \mathcal{O}_o\rangle$, which correspond to the subject, predicate and object respectively, and $p\in\mathcal{P}$.
There could be multiple relationships for a subject-object pair and we represent these predicates as a vector $\mathbf{p}\in \{0,1\}^{|\mathcal{P}|}$, where $\mathbf{p}[i]=1$ indicates the appearance of the $i$-th predicate in $\mathcal{P}$ and the corresponding relationship triplet  $\langle \mathcal{O}_s, \mathcal{P}_i, \mathcal{O}_o\rangle$.
Furthermore, we denote the distributions of the classes and predicates as $\tilde{\mathbf{c}}\in [0,1]^{|\mathcal{C}|}$ and $\tilde{\mathbf{p}}\in[0,1]^{|\mathcal{P}|}$, where  $\sum_i \tilde{\mathbf{c}}[i]=1$.

\section{Methodology}
In this section, we first identify the main challenges in modeling the temporal dynamics, then we discuss how DSG-DETR addresses them. 

\begin{figure*}[htp]
    \centering
    \includegraphics[width=0.95\textwidth]{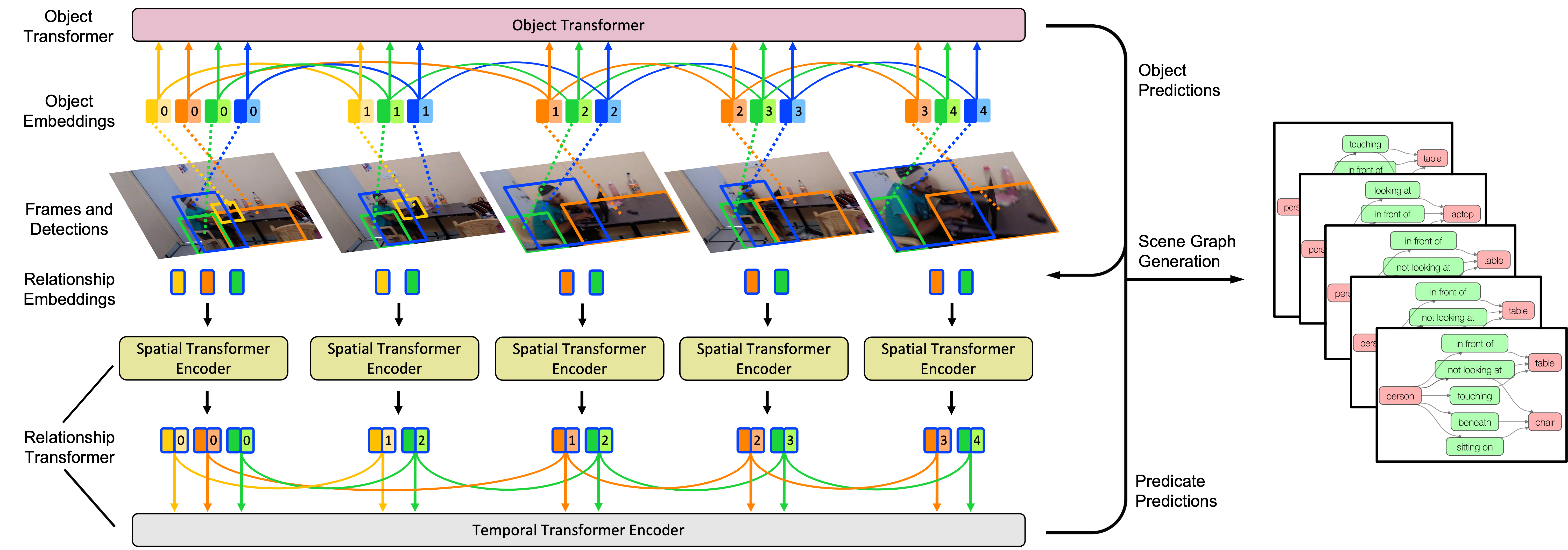}
    \caption{Visualization of the DSG-DETR model. The upper half visualizes the \textit{object transformer} and the lower half sketches the \textit{spatio-temporal relationship transformer}. Each color (orange, tangerine, green and blue) corresponds to one object. The token boxes without borderline represent object embeddings, and those with the borderline represent relationship embeddings, where the colors of the borderline and token box indicate the subject and object respectively.  The token boxes with numbers inside are the positional encoding, e.g., the blue box with ``1" inside 
    denotes its position in the blue tracklet.
    The solid lines connecting the token boxes stand for the tracking results.
    }
    \label{fig:model}
    \vspace{-10pt}
\end{figure*}

\subsection{Temporal dynamics}
Dynamic SGG requires reasoning over both spatial and temporal information. 
However, existing literature lacks a methodical analysis on what kind of information is needed for the temporal and spatial consistencies. 
In our experiments, we find the following aspects  the main challenges for dynamic SGG. \vspace{-5pt}

\paragraph{Temporal object consistency.} 
Static image based object detectors fail to detect video objects consistently due to factors like motion blur, fast movement, occlusion, compression artifacts, and temporal variation of predictions.  
Challenging cases like severe occlusion pose great difficulty in identifying an object from a single frame. 
We show that grounding the predictions over a long-term temporal context and enforcing it to be temporally consistent - i.e., avoiding sudden appearances or disappearances of object representations - results in more accurate and consistent object detections in video.


\vspace{-5pt}
\paragraph{Temporal relationship transition.} 
Besides the temporal object consistency, the other challenge for dynamic SGG is the temporal relation transition. 
Modeling relation transition allows different relationships among the same object-pair over time. 
We aim to maximize the conditional probability of a relationship given the previous relationships and the current observation.

Fig~\ref{fig:model} describes the overall framework of DSG-DETR, which consists of an object transformer and a spatio-temporal relationship transformer, addressing the object consistency and relationship transition respectively.

\subsection{Temporal object consistency}
\subsubsection{Online tracklet construction.}
Unlike \cite{UnifiedGraphStructured2021} that connects all objects in neighboring frames in a sliding window fashion, we only connect the objects which exhibit \emph{apparent} similarity in either the visual feature or the spatial location. Our relatively sparser connections allow us to reason over long-term temporal contexts for a given computational and memory budget.
To this end, we ground our method on a coarse tracking algorithm. 
It is worth noting that the purpose of our tracking is to \textbf{make the transformer only attend to the relevant features efficiently rather than directly extract the correct tracklets}, which is a significant difference between DSG-DETR and previous tracking based methods. 

Prior to the construction of the tracklets, we pass all the frames in a video to Faster R-CNN \cite{ren2016faster} to obtain the object bounding boxes (if not available), object class distributions and object features. We use  $\langle\mathbf{b},\tilde{\mathbf{c}},\mathbf{f}\rangle$ to denote detection, where $\mathbf{b}$ and $\tilde{\mathbf{c}}$ correspond to the bounding box and class distribution respectively and $\mathbf{f}\in\mathbb{R}^{2048}$ is the visual feature vector of the bounding box from Faster R-CNN.

Starting from the first frame, we iteratively match the detections with previous tracklets \cite{he2021gmtracker}, where the tracking results in $i$-th frame could be represented as a permutation $\sigma_i(\cdot)$, for example, $\sigma_i(j)$ assigns the $j$-th detection in the $i$-th frame to $\sigma_i(j)$-th tracklet.

We denote the $j$-th detection in the $i$-th frame as $\mathcal{D}_{ij}=\langle\mathbf{b}_{ij}, \tilde{\mathbf{c}}_{ij}, \mathbf{f}_{ij}\rangle$, and the set of  detections in the $i$-th frame as $\mathcal{D}_i = \{\mathcal{D}_{ij}|j\}$. 
We refer to the $k$-th tracklet up to the $i$-th frame as a set $\mathcal{T}_{(i-1)k}=\{\mathcal{D}_{i'j}| i'\leq i-1, j\leq|\mathcal{D}_{i'}|, \sigma_{i'}(j)=k\}$, which consists of all the detections matched to this tracklet in the previous frames. The set of tracklets up to the $i$-th frame is denoted as $\mathcal{T}_{i-1}=\{\mathcal{T}_{(i-1)k}|k\}$.

For each tracklet $\mathcal{T}_{(i-1)k}$, we define its position $\hat{\mathbf{b}}_{(i-1)k}$ as the bounding box of the last added detection, its class distribution and features as the average of the distributions and features over all detections in it: 
\begin{align}
    \hat{\mathbf{c}}_{(i-1)k} &= \frac{1}{|\mathcal{T}_{(i-1)k}|}\sum_{i'\leq i-1,j\leq|\mathcal{D}_{i'}|} \mathbbm{1}_{[\sigma_{i'}(j)=k]}\tilde{\mathbf{c}}_{i'j}\\
    \hat{\mathbf{f}}_{(i-1)k} &= \frac{1}{|\mathcal{T}_{(i-1)k}|}\sum_{i'\leq i-1,j\leq|\mathcal{D}_{i'}|} \mathbbm{1}_{[\sigma_{i'}(j)=k]}\mathbf{f}_{i'j},
\end{align}
where $\mathbbm{1}$ outputs  $1$ if the condition holds and $0$ otherwise.

A tracklet is regarded as inactive if it does not get any new detection added in the past $m$ frames. 
Here, $m$ is the number of frames which corresponds to the time interval in the original Charades
\cite{sigurdsson2016hollywood} 
videos which were annotated with scene graphs in Action Genome~\cite{Action_genome_Ji_2020_CVPR} dataset.

At the $i$-th frame, the detections of $\mathcal{D}_i$ are only matched with the active tracklets in $\mathcal{T}_{i-1}$. 
Let $n_i=\max\{|\mathcal{D}_i|, |\mathcal{T}_{i-1}|\}$,  we pad with $\emptyset$ (empty set) for detections as $\mathcal{D}'_i=\{\mathcal{D}'_{ij}|j\leq n_i, \mathcal{D}'_{ij}=\mathcal{D}_{ij} \ \text{if}\ j\leq |\mathcal{D}_i|\ \text{else}\ \emptyset \}$ and active tracklets as $\mathcal{T}'_{i-1}=\{\mathcal{T}'_{(i-1)j}|j\leq n_i, \mathcal{T}'_{(i-1)j}=\mathcal{T}_{(i-1)j} \ \text{if}\ j\leq |\mathcal{T}_{i-1}|\ \text{and}\ \mathcal{T}_{(i-1)j}\ \text{is active else}\ \emptyset \}$. 

We use the Hungarian matching algorithm \cite{Kuhn55thehungarian} 
to assign the detections to candidate tracklets 
based on their class distributions, features and positions, which aims to find the permutation of $n_i$ elements $\sigma_i\in\mathfrak{S}_{n_i}$ with the lowest cost, where $\mathfrak{S}_{n_i}$ is the set of all permutations of size $n_i$:
\begin{equation}
\begin{split}
    &\arg\min_{\sigma_i\in\mathfrak{S}_{n_i}}\mathcal{L}_{\text{HM}}(\mathcal{D}'_i, \mathcal{T}'_{i-1})=\\ &\arg\min_{\sigma_i\in\mathfrak{S}_{n_i}}\sum_{j=1}^{|D_i|}\mathbbm{1}_{[\mathcal{T}'_{(i-1)\sigma_i(j)}\neq \emptyset]}[\mathcal{L}_{\text{dist}}(\tilde{\mathbf{c}}'_{ij}, \hat{\mathbf{c}}'_{(i-1)\sigma_i(j)})\\
    +& \mathcal{L}_{\text{feat}}(\mathbf{f}'_{ij}, \hat{\mathbf{f}}'_{(i-1)\sigma_i(j)})+\mathcal{L}_{\text{box}}(\mathbf{b}'_{ij}, \hat{\mathbf{b}}'_{(i-1)\sigma_i(j)})],
\label{eq:match}
\end{split}
\end{equation}
where $\mathcal{L}_{\text{dist}}$, $\mathcal{L}_{\text{feat}}$ and $\mathcal{L}_{\text{box}}$ correspond to the 
loss of the class distributions, features and boxes respectively.

For the class distribution and feature losses, we use the cosine cost such that 
\begin{align}
    \mathcal{L}_{\text{dist}}(\tilde{\mathbf{c}}'_{ij}, \hat{\mathbf{c}}'_{(i-1)\sigma_i(j)}) &=   (1-\cos(\tilde{\mathbf{c}}'_{ij}, \hat{\mathbf{c}}'_{(i-1)\sigma_i(j)}))\\
    \mathcal{L}_{\text{feat}}(\mathbf{f}'_{ij}, \hat{\mathbf{f}}'_{(i-1)\sigma_i(j)}) &= 
  \lambda_{\text{feat}}(1-\cos(\mathbf{f}'_{ij}, \hat{\mathbf{f}}'_{(i-1)\sigma_i(j)})),
\end{align}
where $\cos(\cdot, \cdot)$ represents the cosine similarity and $\lambda_{\text{feat}}$ is a non-negative scalar controlling the weight  between the two loss components.

Following DETR~\cite{carion2020endtoend}, we combine the $L_1$ loss and the generalized IoU loss~\cite{Rezatofighi_2018_CVPR}, denoted as $\mathcal{L}_{\text{iou}}(\cdot, \cdot)$, for the box loss:
\begin{equation}
\begin{split}
    \mathcal{L}_{\text{box}}(\mathbf{b}'_{ij}, \hat{\mathbf{b}}'_{(i-1)\sigma_i(j)}) = \lambda_{\text{iou}}\mathcal{L}_{\text{iou}}(\mathbf{b}'_{ij}, \hat{\mathbf{b}}'_{(i-1)\sigma_i(j)}) \\
    + \lambda_{L_1}\|\mathbf{b}'_{ij}- \hat{\mathbf{b}}'_{(i-1)\sigma_i(j)})\|_1,
\end{split}
\end{equation}
where $\lambda_{\text{iou}}$ and $\lambda_{L_1}$ control the weights of the generalized IoU loss and $L_1$ loss, respectively.

We create a new tracklet for an unmatched object, e.g., when $\mathcal{T}'_{(i-1)\sigma_i(j)}=\emptyset$. 
The Hungarian matching algorithm will always assign a detection to a tracklet,
but it is not guaranteed that the detection indeed has a matching tracklet in $\mathcal{T}'_{i-1}$.  
For example, let's assume a case where the active tracklets correspond to two objects \emph{person} and \emph{table}; but the detections correspond to the objects \emph{person} and \emph{sofa}. The matching algorithm will match the \emph{table} to \emph{sofa}, although they are different. To mitigate this incorrect assignment problem, we ignore the matching  
if the cosine similarity between the features and class distributions are both less than a threshold $\tau$.
In such a case, we mark the corresponding tracklet as empty in the padded tracklet set and create a new tracklet for this detection.

Finally, the existing tracklets in $\mathcal{T}_{i-1}$ can be updated as
\begin{equation}
\label{eq:update1}
\mathcal{T}_{ik} = \mathcal{T}_{(i-1)k} \bigcup \{\mathcal{D}_{ij}|\sigma_i(j)=k, \mathcal{D}'_{ij}\neq \emptyset, \mathcal{T}'_{(i-1)k}\neq \emptyset \}.
\end{equation}

The entire coarse tracking algorithm is summarized in Algorithm \ref{alg:track1}.
\begin{algorithm}[tb]
\caption{Coarse tracking algorithm}
\label{alg:track1}
\begin{algorithmic}[0] 
\STATE \textbf{Input data:} Detections $\mathcal{D}_1, \mathcal{D}_2, \cdots, \mathcal{D}_I$ and video timestamps
\STATE \textbf{Input hyperparameters:} $m$, $\lambda_{\text{feat}}$, $\lambda_{\text{iou}}$, $\lambda_{L_1}$ and $\tau$
\STATE Let $\mathcal{T}_0=\emptyset$
\FOR{iteration  $i=1, 2, \cdots,I$}
\STATE Construct the padded set $\mathcal{D}'_i$ and $\mathcal{T}'_{i-1}$ from $\mathcal{D}_i$ and $\mathcal{T}_{i-1}$
\STATE Compute the optimal matching $\sigma_i$ using Equation \ref{eq:match} 
\STATE $k\leftarrow |\mathcal{T}_{i-1}|+1$
\FOR{iteration $j=1, 2, \cdots, |\mathcal{D}_i|$}
\IF{$\mathcal{T}'_{(i-1)\sigma_j(i)}\neq\emptyset$ and $\cos(\tilde{\mathbf{c}}'_{ij}, \hat{\mathbf{c}}'_{(i-1)\sigma_i(j)})< \tau$ and $ \cos(\tilde{\mathbf{f}}'_{ij}, \hat{\mathbf{f}}'_{(i-1)\sigma_i(j)}) <\tau$}
\STATE $\mathcal{T}'_{(i-1)\sigma_i(j)}\leftarrow\emptyset$
\ENDIF
\IF{$\mathcal{T}'_{(i-1)\sigma_i(j)}=\emptyset$}
\STATE  Create $\mathcal{T}_{ik} \leftarrow \{\mathcal{D}_{ij}\}$ and update $k\leftarrow k+1$
\ENDIF
\ENDFOR
\STATE Update the tracklets in $\mathcal{T}_{i-1}$ according to Equation \ref{eq:update1} 
\STATE Update the tracklet set as $\mathcal{T}_{i}=\{\mathcal{T}_{ik'}|k'=1,2,\cdots, k-1\}$
\ENDFOR
\STATE \textbf{return:} $\mathcal{T}_{I}$
\end{algorithmic}
\end{algorithm}

\subsubsection{Object transformer for long-term consistency.}
We build a transformer on top of these tracklets to realize the temporal object consistency.
For each detection, we represent it as a concatenation of the box embedding, class distribution embedding and object features, which can be written as
\begin{equation}
    \mathbf{o} = \text{Concat}(g^{\text{box}}(\mathbf{b}), g^{\text{dist}}(\tilde{\mathbf{c}}), \mathbf{f}),
\end{equation}
where  $g^{\text{box}}$ and $g^{\text{dist}}$ stand for the embedding functions of the box and class distribution respectively, and $\mathbf{o}\in\mathbb{R}^{d_o}$.

For each tracklet, $\mathcal{T}_{Ik}=\{\mathcal{D}_{ij}| \sigma_i(j)=k\}$, we represent all of its detections  as a matrix $\mathbf{O}_{k}\in\mathbb{R}^{|\mathcal{T}_{Ik}|\times d_o}$.  Then we apply an object transformer with positional encoding $PE(\cdot)$ followed by a feedforward network to output the new object class distributions $\tilde{\mathbf{C}}_k\in [0,1]^{|\mathcal{T}_{Ik}|\times |\mathcal{C}|}$ as:
\begin{align}
    \label{eq:obj_feat}
    \tilde{\mathbf{F}}_k &= \text{Encoder}_{\text{object}}(\mathbf{O}_k + PE(\mathbf{O}_k))\\
    \tilde{\mathbf{C}}_k &= \text{Softmax}(\text{FFN}(\tilde{\mathbf{F}}_k)).
\end{align}
The standard cross entropy loss $\mathcal{L}_{obj}$ is used for the object classification.

\subsection{Temporal relationship transition}
To model the relationship transition, we 
still ground our model on those tracklets, with their predicted subject-object classes, i.e., the relationships sharing the same subject-object classes \footnote{In Action Genome \cite{Action_genome_Ji_2020_CVPR}, each object class is unique in one frame.} across the key frames are in the same sequence.

To simultaneously model the spatial dependency, we first feed all relationships into a spatial encoder, which aggregates the information in each frame, then we apply a 
temporal encoder for the same subject-object pairs across frames. 

The relationships  are  defined over a detected subject-object pair 
$\langle\mathcal{D}_s, \mathcal{D}_o\rangle$.
Similar to STTran~\cite{STTran_2021}, we represent the relationships as a combination of  three embeddings, visual embedding, spatial embedding and semantic embedding:
\begin{align}
    \mathbf{r}^{vs} & =  \text{Concat}(g^s(\tilde{\mathbf{f}}_s), g^o(\tilde{\mathbf{f}}_o))\\
    \mathbf{r}^{sp} & = g^{sp}(\mathbf{u}_{so}\oplus g^{\text{boxes}}(\mathbf{b}_s, \mathbf{b}_o)\\
    \mathbf{r}^{se} & = \text{Concat}(g^{se}(\mathbf{c}_s), g^{se}(\mathbf{c}_o))\\
    \mathbf{r} &= \text{Concat}(\mathbf{r}^{vs},\mathbf{r}^{sp} ,\mathbf{r}^{se} ),
\end{align}
where $\mathbf{r}^{vs}$, $\mathbf{r}^{sp}$, $\mathbf{r}^{se}$ correspond to the visual embedding, spatial embedding and semantic embedding, respectively. $\tilde{\mathbf{f}}$ is the spatial-temporal visual feature computed in Equation \ref{eq:obj_feat}.
$g^s$ and $g^o$ are the visual feature embedding functions for the subject and object. $g^{sp}$ is the spatial embedding function, whose input is the sum of the union feature $\mathbf{u}_{so}$ for the subject and object extracted by ROIAlign \cite{he2018mask} and a boxes embedding encoded by $g^{\text{boxes}}$, where $\oplus$ stands for the element-wise addition. $g^{se}$ is the word embedding of the object class. Please refer to the Supplementary material for more details about each embedding function.

We stack all relationships $\mathbf{r}$ in the $i$-th frame into a matrix $\mathbf{R}_i$. The output of the spatial transformer encoder is:
\begin{equation}
    \mathbf{R}'_{i} = \text{Encoder}_{\text{spatial}}(\mathbf{R}_{i}).
\end{equation}

Then we rearrange all the output relationship representations $\mathbf{r}'$ according to their subject and object classes.  We stack the relationship representations from the spatial encoder with the subject-object classes $\langle s,o \rangle$ into a matrix $\mathbf{R}'_{so}$, then the logits of the predicates are output by the temporal transformer encoder
\begin{equation}
    \mathbf{Z}_{so} = \text{Encoder}_{\text{temporal}}(\mathbf{R}'_{so} +PE(\mathbf{R}'_{so})).
\end{equation}

The predicates logits $\mathbf{z}$ of the corresponding relationship representation $\mathbf{r'}$ go through different linear projections to obtain the final predicates distribution $\tilde{\mathbf{p}}$ for different HOI types belonging to \emph{attention}, \emph{spatial} and \emph{contact} as defined in the Action Genome dataset~\cite{Action_genome_Ji_2020_CVPR}.
We use the multi-label margin loss for the predicate classification, 
\begin{equation}
    \mathcal{L}_p(\tilde{\mathbf{p}}) = \sum_{i \in\mathcal{P}^+}\sum_{j \in \mathcal{P}^-}\max(0, 1-\tilde{\mathbf{p}}[j] +\tilde{\mathbf{p}}[i]),
\end{equation}
where $\mathcal{P}^+$ denotes the indices of the annotated predicates and $\mathcal{P}^-$ denotes the indices of the predicates not in the annotation.
The final loss combines both the object loss and predicate loss $\mathcal{L}_{\text{total}} = \mathcal{L}_{obj}+\mathcal{L}_p$.




\section{Experiments}

\subsection{Experimental setup} \label{questions}


\paragraph{Dataset} We evaluate our method on Action Genome  dataset \cite{Action_genome_Ji_2020_CVPR} containing 35 object categories and 25 relationship categories. The relationships are categorized into three human-object categories: attention, spatial and contact relationships, where multiple relationships may appear in spatial and contact categories. 


\paragraph{Training} We use one NVIDIA Tesla V100S GPU with 32G memory for training. Similar to ~\cite{STTran_2021}, we utilize Faster R-CNN with Resnet101 \cite{he2015deep} backbone for the object detector and pretrain it on Action Genome~\cite{Action_genome_Ji_2020_CVPR}. All layers in the backbone for the object detector feature extraction are frozen when training our method. We use AdamW \cite{loshchilov2019decoupled} to optimize with batch size $1$. The initial learning rate is set to $10^{-5}$. 

\paragraph{Evaluation} The edge (predicate) classification conditioned on nodes (subject \& object) is a relatively easy problem to solve. Most methods reach the ballpark of 99\% in the unconstrained Recall@$K$=50 for this \textit{PredCls} task. 
The problem becomes challenging when the joint object classification / detection is involved  in SGCls / SGDet tasks. We evaluate performances on these two tasks. 
(1) scene graph classification \underline{(SGCls)}: where the video frames and bounding boxes are provided and the task is to predict the predicates and subject/object classes.
(2) scene graph detection \underline{(SGDet)}: The task is to detect the objects and predict the predicates for object pairs, where only the video frames are provided.  
Following the convention of object detection, in case of SGDet, an entity (subject or object) is regarded as successfully detected if the Intersection-Over-Union (IOU) between the predicted bounding box and the ground-truth bounding box is larger than $0.5$ and the predicted and ground-truth class labels match.
Please refer to the supplementary material for additional results on PredCls. 
We use Recall@K (R@K, K=[10,20,50]) \cite{lu2016visual,Action_genome_Ji_2020_CVPR} as the evaluation metric, which measures the fraction of the ground-truth relationship triplets in the top K predictions. 


For the relationship tuple of a subject-object pair $\langle \mathcal{D}_s, \mathcal{D}_o\rangle$, we define the score of each detected object as the highest class score in its distribution, $\max\{\tilde{\mathbf{c}}\}$. Then the score of $i$-th relationship triplet is  estimated as the product of three scores:
\begin{equation}
    \max\{\tilde{\mathbf{c}}_s\}\cdot \tilde{\mathbf{p}}[i]\cdot \max\{\tilde{\mathbf{c}}_o\}.
\end{equation}

For the calculation of R@K, the relationship triplets are ordered according to their scores among all relationship triplets of that category in a frame.

\begin{table*}[tb!]
\centering
\caption{Comparison with state-of-the-art scene graph generation methods on Action Genome~\cite{Action_genome_Ji_2020_CVPR}. 
Note that $\star$ denotes results reproduced from the official model~\cite{TRACE} for the same evaluation setup as others.
}
\setlength{\tabcolsep}{2pt}
\resizebox{1\textwidth}{!}{%
\begin{tabular}{c ccccccccc ccccccccc}
\toprule
  & \multicolumn{6}{c}{With Constraint} & \multicolumn{9}{c}{No Constraints}\\
  \cmidrule(lr){2-7} \cmidrule(lr){8-13}
  Method & 
   \multicolumn{3}{c}{SGCls} & \multicolumn{3}{c}{SGDet} &  \multicolumn{3}{c}{SGCls} & \multicolumn{3}{c}{SGDet} \\ \cmidrule(lr){2-4} \cmidrule(lr){5-7} \cmidrule(lr){8-10} \cmidrule(lr){11-13} \cmidrule(lr){14-16}
   & R@10 & R@20 & R@50 & R@10 & R@20 & R@50 & R@10 & R@20 & R@50  & R@10 & R@20 & R@50
\\
\midrule
\midrule
VRD~\cite{lu2016visual}  & 32.4   &33.3 &33.3  &19.2 &24.5 &26.0  &39.2 &49.8 &52.6 &19.1 &28.8 &40.5\\
M-FREQ~\cite{neural_motifs_2017}    &40.8 &41.9 &41.9 &23.7 &31.4 &33.3  &50.4 &60.6 &64.2 &22.8 &34.3 &46.4 \\
MSDN~\cite{Li2017SceneGG}  &43.9 &45.1 &45.1 &24.1 &32.4 &34.5 &51.2& 61.8 &65.0 &23.1 &34.7 &46.5 \\
VCTree~\cite{VCTree_Tang_2019_CVPR}  &44.1 &45.3 &45.3 &24.4 &32.6 &34.7  &52.4 &62.0 &65.1 &23.9 &35.3 &46.8 \\
RelDN~\cite{zhang2019vrd}  &44.3 &45.4 &45.4 &24.5 &32.8 &34.9  &52.9 &62.4 &65.1 &24.1 &35.4 &46.8  \\
GBS-Net~\cite{lin2020gps}  &45.3 &46.5 &46.5 &24.7 &33.1 &35.1  &53.6 &63.3 &66.0 &24.4 &35.7 &47.3  \\
TRACE~\cite{TRACE}$\star$  &14.8 &14.8 &14.8 &13.9 &14.5 &14.6 &37.1 & 46.7 &50.5 &26.5 &35.1 &45.3\\ 
STTran~\cite{STTran_2021} &46.4 &47.5 &47.5 &25.2 &34.1 &37.0  &54.0 &63.7 &66.4 &24.6 &36.2 &48.8 \\
APT~\cite{Li_2022_CVPR} &47.2 &48.9 &48.9 &26.3 &\bf 36.1  &\bf 38.3 &55.1 &65.1 &68.7 &25.7 &37.9 &\bf 50.1 \\

{\bf DSG-DETR(Ours)} &\textbf{50.8} &\textbf{52.0} &\textbf{52.0} 
&\textbf{30.3} &34.8 & 36.1
&\textbf{59.2} &\textbf{69.1} & \textbf{72.4} 
&\textbf{32.1} & \textbf{40.9}& 48.3 \\
\bottomrule
\end{tabular}
\label{tab:main_table}}
\vspace{-15pt}
\end{table*}

\subsection{Comparison with SOTA 
}

Table \ref{tab:main_table} shows the main result of the proposed DSG-DETR.
We use STTran \cite{STTran_2021} as our one of the strongest baselines, and develop our DSG-DETR atop their source code~\cite{STTran_code}. 
Besides, we also select some powerful scene graph generation methods on the static images such as VRD \cite{lu2016visual}, M-FREQ \cite{neural_motifs_2017}, MSDN \cite{VCTree_Tang_2019_CVPR}, RelDN \cite{zhang2019vrd} and GBS-Net \cite{lin2020gps}. 
For a fair comparison, we use the same object detector, a pretrained Faster R-CNN fine-tuned on Action Genome for all the baselines. 
Please note that the results for TRACE ~\cite{TRACE} correspond to the same evaluation criteria as with others; in their original paper the setup was different. 
The results show that DSG-DETR outperforms the strongest baseline of STTran in both SGCls and SGDet tasks where long-term dependencies are essential for consistent object recognition. 
DSG-DETR clearly outperforms the state-of-art by $\sim$10\% and $\sim$20\%-30\% in terms of R@10 under constraint and no constraint criteria for SGCls and SGDet, respectively. 
The long-term dependencies in DSG-DETR brings significant improvement.



We observe that the improvement of DSG-DETR becomes less significant when it comes to larger $K$ for SGDet, this is in fact a tradeoff between the consistency and the diversity. Please note, Recall at lower values of $K$ are more significant than the larger values of $K$ when such models are expected to be used for downstream applications. 

\subsection{Object-level consistency on object-tracks
}
To study the best possible effect of modeling object-level consistency, 
we apply the same framework on ground-truth object tracks instead of the constructed tracklets. 
Such sequences can be treated as the hypothetical \emph{best case} for exploiting such long-term dependencies. 
We see in Table \ref{tab:ub_table} that the \emph{best case} scenario of exploiting the long-term dependencies largely improves the performance over the baseline; Specifically, SGCls improves by $\sim$13\%-14\% and SGDet improves by $\sim$ 45\%-46\% over the baseline. 
We also show that our proposed tracking algorithm in DSG-DETR helps in reducing the gap between the baseline and the ground-truth tracklet upper-bound in all cases.  
For example, the \emph{best case} for SGCls performs only $\sim$4\% better than DSG-DETR in terms of R@10.
For SGDet, DSG-DETR is able to to reduce the performance gap from the \emph{best case} by almost half comparing with the baseline 
($\sim$45\% vs $\sim$21\%) in terms of R@10. 

\begin{table*}[tb!]
\centering
\caption{Hypothetical \emph{best case} when exploiting long-term dependencies captured via long-term object tracks. 
GTTrack includes all components of DSG-DETR but uses the ground-truth object tracks instead. DSG-DETR relies on online tracklet construction. DSG-DETR outperforms the baseline  
and significantly minimizes the performance gap between baseline and GTTrack.  
}
\setlength{\tabcolsep}{12pt}
\resizebox{1\linewidth}{!}{%
\begin{tabular}{c cccc cccc}
\toprule
  & \multicolumn{4}{c}{With Constraint} & \multicolumn{4}{c}{No Constraints}\\
  \cmidrule(lr){2-5} \cmidrule(lr){6-9}
  Method & 
  \multicolumn{2}{c}{SGCls} & \multicolumn{2}{c}{SGDet} & \multicolumn{2}{c}{SGCls} & \multicolumn{2}{c}{SGDet} \\ \cmidrule(lr){2-3} \cmidrule(lr){4-5} \cmidrule(lr){6-7} \cmidrule(lr){8-9} 
  & R@10 & R@20 & R@10 & R@20& R@10 & R@20 & R@10 & R@20
\\
\midrule
\midrule
Baseline(STTran~\cite{STTran_code}) & 45.7 & 46.8 & 25.2 & 34.1 & 54.2 & 63.5 & 24.6 & 36.2 \\
GTTrack (Upper-bound) &\bf 52.2 & \bf 53.4 & \textbf{36.8} &\textbf{37.9} & \bf 60.9& \bf 71.0 & \bf 43.8& \bf 51.0 \\
\midrule
DSG-DETR(Ours) & 50.8 & 52.0 & 30.3 & 34.8 & 59.2 & 69.1 & 32.1 & 40.9 \\
\bottomrule
\end{tabular}
\label{tab:ub_table}}
\vspace{-15pt}
\end{table*}


\begin{table}[tb]
\caption{Ablation on different components of DSG-DETR for SGCls on Action Genome. 
}
\label{tab:ablation}
\centering
\vspace{-2mm}
\setlength{\tabcolsep}{1pt}
\resizebox{1\linewidth}{!}{%
\begin{tabular}{ccc cc cc}
\toprule
{Obj-trans} & {Pos-enc} & {Rela-trans}  &  \multicolumn{2}{c}{With Constraint} &\multicolumn{2}{c}{No Constraint} \\
\cmidrule(lr){4-5} \cmidrule(lr){6-7}
& &  & {R@20} & {R@50} & {R@20} & {R@50}\\ 
\midrule
 -& -&  -&  46.8 & 46.8 & 63.5 & 66.0       \\
-& -& \checkmark &  47.1 & 47.1 &63.5&65.9 \\
\checkmark &  \checkmark & -& 51.4 & 51.4  &68.7 & 71.9            \\
\checkmark & \checkmark &  \checkmark & \textbf{52.0} &\textbf{52.0} &\textbf{69.1} &\textbf{72.4} \\
\bottomrule
\end{tabular}
}
\end{table}


\begin{table}[tb]
\caption{Effect of temporal context on SGCls. Short-term context corresponds to  tracklet construction over 5 key-frames only, whereas the long-term context uses an order of magnitude higher number of key-frames (usually 5 to 40 key-frames), estimated by our tracklet construction algorithm.  
}
\label{tab:ablation_temporal_context}
\centering
\vspace{-2mm}
\setlength{\tabcolsep}{1pt}
\resizebox{1\linewidth}{!}{%
\begin{tabular}{c ccc ccc c}
\toprule
Time context & \multicolumn{3}{c}{With Constraint} &\multicolumn{3}{c}{No Constraint} & Obj Acc\\
\cmidrule(lr){2-4} \cmidrule(lr){5-7}
& {R@10} & {R@20} & {R@50} & {R@10} & {R@20} & {R@50}  \\ 
\midrule
no-context  &  45.7  & 46.8 & 46.8 & 54.2 & 63.5 & 66.0   &   70.2 \\
short-term & 47.4 & 48.6 & 48.6 &55.6 &65.3 &68.1  & 73.0\\
long-term &\textbf{50.8} &\textbf{52.0} &\textbf{52.0} &\textbf{59.2} &\textbf{69.1} &\textbf{72.4} & \textbf{73.8}\\
\bottomrule
\end{tabular}
}
\vspace{-10pt}
\end{table}

\subsection{Ablation studies}
In DSG-DETR, we propose to capture long-term dependencies primarily by consistent and effective object tracklets construction. We employ two transformers - one for consistent object prediction and the other for relationship transitions. For the object transformer, we additionally integrate the temporal position with the object representations 
into the transformer encoder 
through positional encoding, denoted by ``Pos enc" in the ablation Table \ref{tab:ablation}. 
We use the sinusoidal encoding from~\cite{vaswani2017attention}.
Our relationship transformer architecture shares the same spatial encoder as STTran~\cite{STTran_2021}, but it replaces the temporal decoder in STTran with a temporal encoder operating on the predicted classes sequences for capturing the long-term dependencies. In Table \ref{tab:ablation}, we replace our relationship transformer with STTran for ablation. We show the results of ablating our model for SGCls task. 

The heavy lifting is done by the object transformer employed on the constructed tracklets. 
For relationship transformer, the first two rows in the table demonstrate that even with the predicted classes sequences based on Faster R-CNN results, capturing the long-term dependencies still brings the benefits with 0.3 point improvement compared with STTran in R@K under constraint. Finally,
the positional encoding in the object transformer boosts the performance by additional 0.2 point in terms of R@K under constraint and even significant improvement for no constraint evaluation.
The ablative studies for SGDet also exhibits similar trend, and available in the Supplementary material.  

Table ~\ref{tab:ablation_temporal_context} shows how long-term temporal context helps improve the video object detection and dynamic SGG performances. 
Long-term temporal context significantly improves SGG by ~3-6 points over the baseline (row 3 vs row 1).
For short-term temporal context, the performance drops by ~3-4 points (row 2 vs row 3), yet beating the baseline. It also shows how object classification accuracy improves as we move from no-temporal context to short-term context to long-term context. 
Object classification accuracy significantly contributes to the fraction of correct triplet predictions as measured by Recall.

\begin{figure}
\centering
  \begin{subfigure}{0.3\linewidth}
    \includegraphics[width=\linewidth]{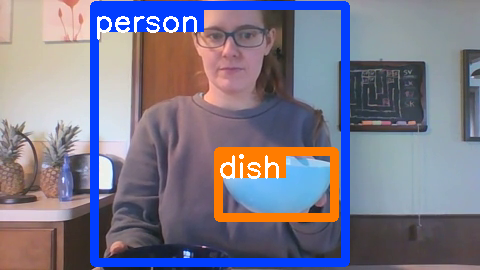}
  \caption{Correctly classified by all the models.}
  \end{subfigure}
  \begin{subfigure}{0.3\linewidth}
    \includegraphics[width=\linewidth]{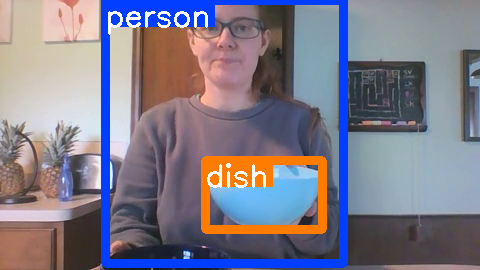}
  \caption{DSG-DETR correctly predicts 'dish' objects.}
  \end{subfigure}
  \begin{subfigure}{0.3\linewidth}
    \includegraphics[width=\linewidth]{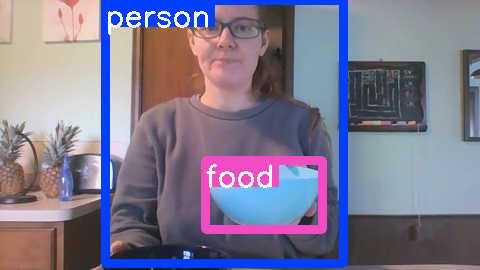}
  \caption{STTran (F-RCNN) misclassifies 'dish' as 'food'.}
  \end{subfigure}
  \caption{DSG-DETER predicts temporally consistent objects }
  \label{fig:temp1}
  \vspace{-10pt}
\end{figure}

\begin{figure}
\centering
    \begin{subfigure}{0.4\linewidth}
    \centering
    \includegraphics[width=0.5\linewidth]{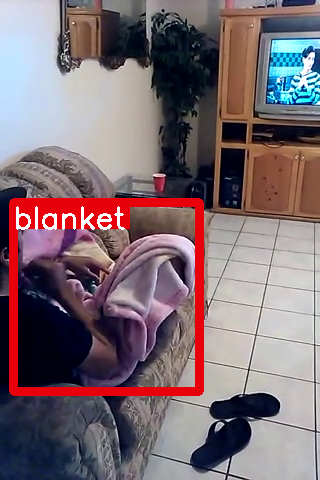}
    \caption{Correct object prediction.}
    \end{subfigure}
    \hspace{20pt}
    \begin{subfigure}{0.4\linewidth}
    \includegraphics[width=0.9\linewidth]{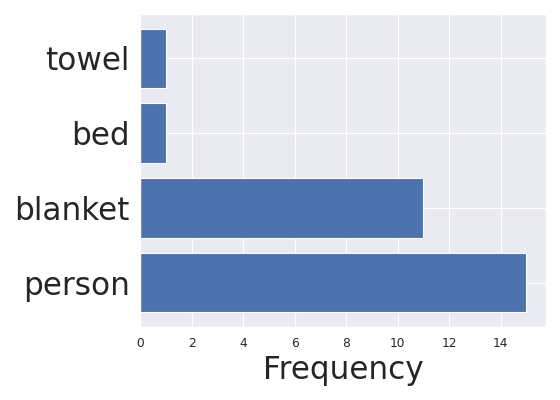}
    \caption{Histogram of Faster R-CNN predictions on the \emph{blanket} object in the tracklet.}
    \end{subfigure}
    \caption{DSG-DETR recovers from the F-RCNN mis-classification 
    leveraging context from the tracklet.
    }
    \label{fig:temp2}
\vspace{-15pt}
\end{figure}

\subsection{Qualitative results}
Fig \ref{fig:temp1} is an example where DSG-DETR successfully constructs the sequence of the blue bowl (Figs \ref{fig:temp1}(a) and \ref{fig:temp1}(b)) from the temporally ordered key frames (top to bottom) and makes the correct prediction ``dish'' for all of them. While Faster R-CNN and STTran will mis-classify as shown in Fig \ref{fig:temp1}(c). 
Fig \ref{fig:temp2} shows an example frame where DSG-DETR constructed a tracklet of a blanket shown in the red bounding box.
Most of the detections in the tracklet are predicted as ``person" by Faster R-CNN shown in Fig \ref{fig:temp2} (b).
However, the object transformer makes a correct prediction ``blanket" for all of them in the sequence. 
This reveals that the object transformer in fact learns to reason over temporal dependencies beyond a simple majority voting.



\begin{figure}
    \centering
    \includegraphics[trim={2mm 1mm 1mm 1mm},clip, width=0.7\linewidth]{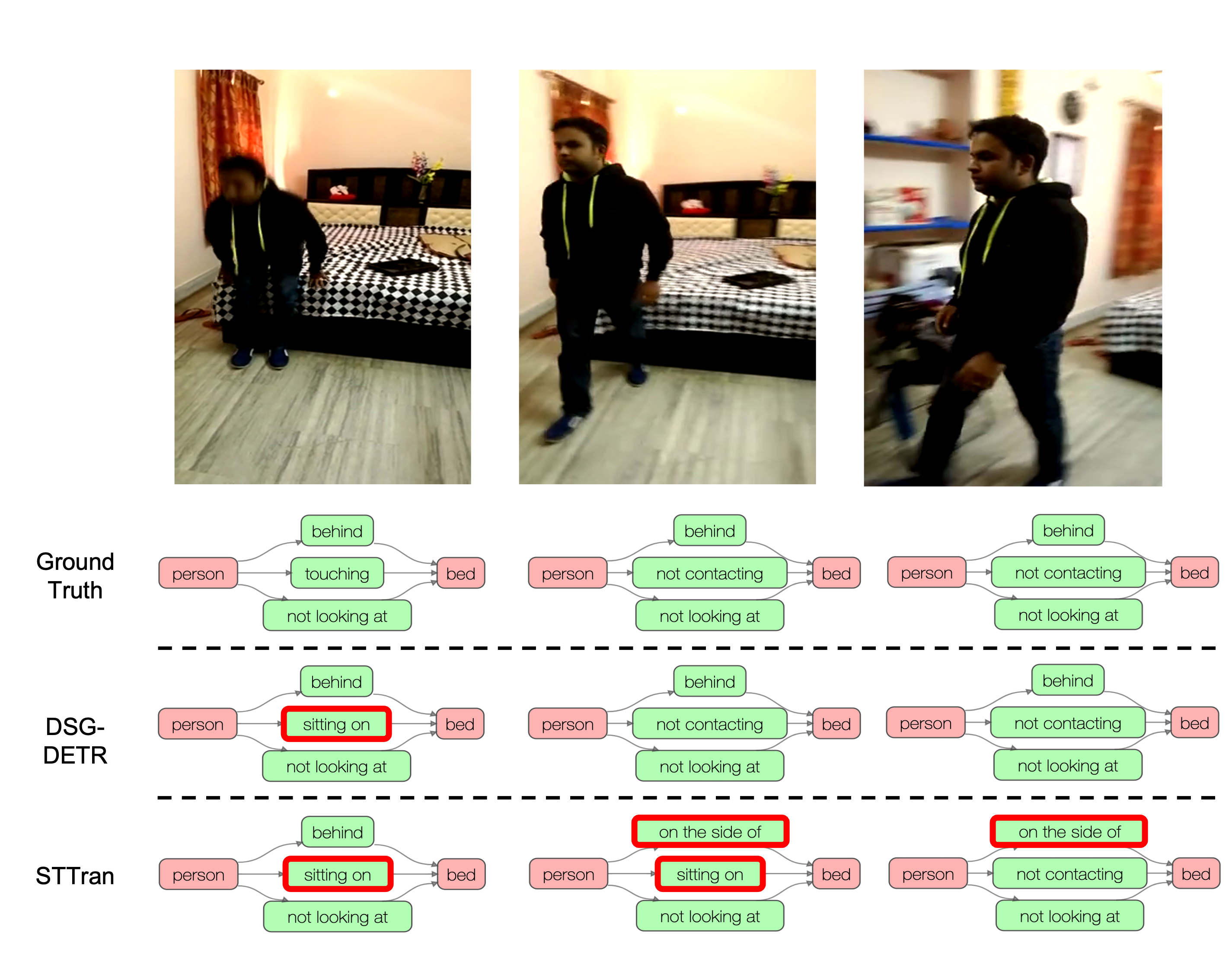}
    \caption{Scene graphs generated by DSG-DETR and STTran for three key frames sampled from an action which embeds long-term dependencies.
    }
    \label{fig:sg}
    \vspace{-10pt}
\end{figure}

In Fig \ref{fig:sg}, we sample three key frames for an action where a person gets up from the bed and walks towards the doorway. The first frame is 
33 frames 
away from the third one in the original video. 
Thanks to its capability to exploit long-term dependencies, DSG-DETR successfully understands the whole action with only one mistake which is mis-classifying ``touching" as ``sitting on" in the first frame. 
However, STTran makes many mistakes including predicting that the human is still ``sitting on" the bed in the second frame and the bed is ``on the side of" the human in the third frame rather than "behind". 

\section{Conclusions}

We hypothesized that capturing long-term temporal context is crucial for dynamic scene graph generation. 
We presented a framework called Dynamic Scene Graph Detection Transformer (DSG-DETR) that is capable of exploiting long-term dependencies within object-tracklets constructed in an online fashion. 
We also estimated an upper-bound on the performance of dynamic SGG tasks leveraging such notion of long-term consistencies by utilizing the ground-truth object tracks, and show that DSG-DETR is able to noticeably minimize the performance gap between this upper-bound and the baseline. 
We demonstrated the efficacy of DSG-DETR on the Action Genome dataset, where it significantly outperforms the state-of-the-art methods.

{\small
\bibliographystyle{ieee_fullname}
\bibliography{main}

\begin{thebibliography}{10}\itemsep=-1pt

\bibitem{STTran_code}
Spatial-temporal transformer for dynamic scene graph generation.
\newblock \url{https://urldefense.com/v3/__https://github.com/yrcong/STTran__;!!DZ3fjg!8acGTt-4Bh7ERSxc5HZloo9DB6erQ6aUZCU4Aeyn-Hz19s1YanR5hh-VwkpjPlzn1UErL3rEiPUNddBvcMaO-kLQ9wgqQyk$  }.
\newblock Accessed: 2021-08-30.

\bibitem{UnifiedGraphStructured2021}
Anurag Arnab, Chen Sun, and Cordelia Schmid.
\newblock Unified {{Graph Structured Models}} for {{Video Understanding}}.
\newblock In {\em Proceedings of the IEEE/CVF International Conference on
  Computer Vision (ICCV)}, 2021.

\bibitem{DETRCarionMSUKZ20}
Nicolas Carion, Francisco Massa, Gabriel Synnaeve, Nicolas Usunier, Alexander
  Kirillov, and Sergey Zagoruyko.
\newblock End-to-end object detection with transformers.
\newblock In Andrea Vedaldi, Horst Bischof, Thomas Brox, and Jan{-}Michael
  Frahm, editors, {\em Computer Vision - {ECCV} 2020 - 16th European
  Conference, Glasgow, UK, August 23-28, 2020, Proceedings, Part {I}}, volume
  12346 of {\em Lecture Notes in Computer Science}, pages 213--229. Springer,
  2020.

\bibitem{carion2020endtoend}
Nicolas Carion, Francisco Massa, Gabriel Synnaeve, Nicolas Usunier, Alexander
  Kirillov, and Sergey Zagoruyko.
\newblock End-to-end object detection with transformers, 2020.

\bibitem{STTran_2021}
Yuren Cong, Wentong Liao, Hanno Ackermann, Bodo Rosenhahn, and Michael~Ying
  Yang.
\newblock Spatial-{{Temporal Transformer}} for {{Dynamic Scene Graph
  Generation}}.
\newblock In {\em Proceedings of the International Conference on Computer
  Vision (ICCV)}, October 2021.

\bibitem{dosovitskiy2020vit}
Alexey Dosovitskiy, Lucas Beyer, Alexander Kolesnikov, Dirk Weissenborn,
  Xiaohua Zhai, Thomas Unterthiner, Mostafa Dehghani, Matthias Minderer, Georg
  Heigold, Sylvain Gelly, Jakob Uszkoreit, and Neil Houlsby.
\newblock An image is worth 16x16 words: Transformers for image recognition at
  scale.
\newblock {\em ICLR}, 2021.

\bibitem{Gao2021}
Kaifeng Gao, Long Chen, Yifeng Huang, and Jun Xiao.
\newblock Video relation detection via tracklet based visual transformer.
\newblock {\em Proceedings of the 29th ACM International Conference on
  Multimedia}, Oct 2021.

\bibitem{eccv_GarciaN20}
Noa Garcia and Yuta Nakashima.
\newblock Knowledge-based video question answering with unsupervised scene
  descriptions.
\newblock In {\em {ECCV} {(18)}}, volume 12363 of {\em Lecture Notes in
  Computer Science}, pages 581--598. Springer, 2020.

\bibitem{GirdharCDZ19}
Rohit Girdhar, Jo{\~{a}}o Carreira, Carl Doersch, and Andrew Zisserman.
\newblock Video action transformer network.
\newblock In {\em {CVPR}}, pages 244--253. Computer Vision Foundation / {IEEE},
  2019.

\bibitem{he2021gmtracker}
Jiawei He, Zehao Huang, Naiyan Wang, and Zhaoxiang Zhang.
\newblock Learnable graph matching: Incorporating graph partitioning with deep
  feature learning for multiple object tracking.
\newblock In {\em Proceedings of the IEEE/CVF Conference on Computer Vision and
  Pattern Recognition (CVPR)}, pages 5299--5309, June 2021.

\bibitem{he2018mask}
Kaiming He, Georgia Gkioxari, Piotr Dollár, and Ross Girshick.
\newblock Mask r-cnn, 2018.

\bibitem{he2015deep}
Kaiming He, Xiangyu Zhang, Shaoqing Ren, and Jian Sun.
\newblock Deep residual learning for image recognition, 2015.

\bibitem{HOI_video_2021}
Jingwei Ji, Rishi Desai, and Juan~Carlos Niebles.
\newblock Detecting human-object relationships in videos.
\newblock In {\em Proceedings of the International Conference on Computer
  Vision (ICCV)}, October 2021.

\bibitem{Action_genome_Ji_2020_CVPR}
Jingwei Ji, Ranjay Krishna, Li Fei-Fei, and Juan~Carlos Niebles.
\newblock Action {{Genome}}: {{Actions}} as {{Composition}} of
  {{Spatio}}-temporal {{Scene Graphs}}.
\newblock In {\em Proceedings of the IEEE/CVF Conference on Computer Vision and
  Pattern Recognition (CVPR)}, June 2020.

\bibitem{image_retrieval_SG_2015}
Justin Johnson, Ranjay Krishna, Michael Stark, Li-Jia Li, David~A. Shamma,
  Michael~S. Bernstein, and Li Fei-Fei.
\newblock Image retrieval using scene graphs.
\newblock In {\em 2015 IEEE Conference on Computer Vision and Pattern
  Recognition (CVPR)}, pages 3668--3678, 2015.

\bibitem{Jung2021}
Gayoung Jung, Jonghun Lee, and Incheol Kim.
\newblock Tracklet pair proposal and context reasoning for video scene graph
  generation.
\newblock {\em Sensors}, 21(9), 2021.

\bibitem{krishnavisualgenome}
Ranjay Krishna, Yuke Zhu, Oliver Groth, Justin Johnson, Kenji Hata, Joshua
  Kravitz, Stephanie Chen, Yannis Kalantidis, Li-Jia Li, David~A. Shamma,
  Michael~S. Bernstein, and Li Fei-Fei.
\newblock Visual genome: Connecting language and vision using crowdsourced
  dense image annotations.
\newblock {\em Int. J. Comput. Vision}, 123(1):32–73, May 2017.

\bibitem{Kuhn55thehungarian}
H.~W. Kuhn and Bryn Yaw.
\newblock The hungarian method for the assignment problem.
\newblock {\em Naval Res. Logist. Quart}, pages 83--97, 1955.

\bibitem{Li2017SceneGG}
Yikang Li, Wanli Ouyang, Bolei Zhou, Kun Wang, and Xiaogang Wang.
\newblock Scene graph generation from objects, phrases and region captions.
\newblock {\em 2017 IEEE International Conference on Computer Vision (ICCV)},
  pages 1270--1279, 2017.

\bibitem{Li_2022_CVPR}
Yiming Li, Xiaoshan Yang, and Changsheng Xu.
\newblock Dynamic scene graph generation via anticipatory pre-training.
\newblock In {\em Proceedings of the IEEE/CVF Conference on Computer Vision and
  Pattern Recognition (CVPR)}, pages 13874--13883, June 2022.

\bibitem{lin2020gps}
Xin Lin, Changxing Ding, Jinquan Zeng, and Dacheng Tao.
\newblock Gps-net: Graph property sensing network for scene graph generation.
\newblock In {\em Proc. IEEE Conf. Comput. Vis. Pattern Recognit.}, pages
  3746--3753, 2020.

\bibitem{loshchilov2019decoupled}
Ilya Loshchilov and Frank Hutter.
\newblock Decoupled weight decay regularization, 2019.

\bibitem{lu2016visual}
Cewu Lu, Ranjay Krishna, Michael Bernstein, and Li Fei-Fei.
\newblock Visual relationship detection with language priors.
\newblock In {\em European Conference on Computer Vision}, 2016.

\bibitem{pmlr_v80_parmar18a}
Niki Parmar, Ashish Vaswani, Jakob Uszkoreit, Lukasz Kaiser, Noam Shazeer,
  Alexander Ku, and Dustin Tran.
\newblock Image transformer.
\newblock In Jennifer Dy and Andreas Krause, editors, {\em Proceedings of the
  35th International Conference on Machine Learning}, volume~80 of {\em
  Proceedings of Machine Learning Research}, pages 4055--4064. PMLR, 10--15 Jul
  2018.

\bibitem{ren2016faster}
Shaoqing Ren, Kaiming He, Ross Girshick, and Jian Sun.
\newblock Faster r-cnn: Towards real-time object detection with region proposal
  networks, 2016.

\bibitem{Rezatofighi_2018_CVPR}
Hamid Rezatofighi, Nathan Tsoi, JunYoung Gwak, Amir Sadeghian, Ian Reid, and
  Silvio Savarese.
\newblock Generalized intersection over union.
\newblock In {\em The IEEE Conference on Computer Vision and Pattern
  Recognition (CVPR)}, June 2019.

\bibitem{sigurdsson2016hollywood}
Gunnar~A. Sigurdsson, Gül Varol, Xiaolong Wang, Ali Farhadi, Ivan Laptev, and
  Abhinav Gupta.
\newblock Hollywood in homes: Crowdsourcing data collection for activity
  understanding, 2016.

\bibitem{videoBERT_Sun_2019_ICCV}
Chen Sun, Austin Myers, Carl Vondrick, Kevin Murphy, and Cordelia Schmid.
\newblock Videobert: A joint model for video and language representation
  learning.
\newblock In {\em Proceedings of the IEEE/CVF International Conference on
  Computer Vision (ICCV)}, October 2019.

\bibitem{VCTree_Tang_2019_CVPR}
Kaihua Tang, Hanwang Zhang, Baoyuan Wu, Wenhan Luo, and Wei Liu.
\newblock Learning to compose dynamic tree structures for visual contexts.
\newblock In {\em The IEEE Conference on Computer Vision and Pattern
  Recognition (CVPR)}, June 2019.

\bibitem{TargetAdaptiveContext_2021}
Yao Teng, Limin Wang, Zhifeng Li, and Gangshan Wu.
\newblock Target {{Adaptive Context Aggregation}} for {{Video Scene Graph
  Generation}}.
\newblock In {\em Proceedings of the International Conference on Computer
  Vision (ICCV)}, October 2021.

\bibitem{TRACE}
Yao Teng, Limin Wang, Zhifeng Li, and Gangshan Wu.
\newblock Target adaptive context aggregation for video scene graph generation.
\newblock In {\em Proceedings of the IEEE/CVF International Conference on
  Computer Vision}, pages 13688--13697, 2021.

\bibitem{vaswani2017attention}
Ashish Vaswani, Noam Shazeer, Niki Parmar, Jakob Uszkoreit, Llion Jones,
  Aidan~N Gomez, \L~ukasz Kaiser, and Illia Polosukhin.
\newblock Attention is all you need.
\newblock In I. Guyon, U.~V. Luxburg, S. Bengio, H. Wallach, R. Fergus, S.
  Vishwanathan, and R. Garnett, editors, {\em Advances in Neural Information
  Processing Systems}, volume~30. Curran Associates, Inc., 2017.

\bibitem{neural_motifs_2017}
Rowan Zellers, Mark Yatskar, Sam Thomson, and Yejin Choi.
\newblock Neural motifs: Scene graph parsing with global context.
\newblock {\em CoRR}, abs/1711.06640, 2017.

\bibitem{zhang2019vrd}
Ji Zhang, Kevin~J. Shih, Ahmed Elgammal, Andrew Tao, and Bryan Catanzaro.
\newblock Graphical contrastive losses for scene graph parsing.
\newblock In {\em CVPR}, 2019.

\end{thebibliography}
}
\newpage
\appendix
\begin{table}[tb]
\centering
\caption{Ablation experiments for SGDet. The first row denotes the reproduced results of STTran \cite{STTran_2021}, with its relationship transformer.  
}
\label{tab:ablation1}
\setlength{\tabcolsep}{2pt}
\resizebox{0.95\linewidth}{!}{%
\begin{tabular}{ccc ccc cccc}
\toprule
{Obj-tr} & {Pos-enc} & Rela-tr & \multicolumn{3}{c}{With Constraint} &\multicolumn{3}{c}{No Constraint} & Obj mAP\\
\cmidrule(lr){4-6} \cmidrule(lr){7-9}
& & & {R@10} & {R@20} & {R@50} & {R@10} & {R@20} & {R@50}\\ 
\midrule
 -& -&   -&  25.1  & 33.9 & 36.8 & 24.6 & 36.2 & 48.9 & 11.4     \\
-& -& \checkmark & 25.3 & 34.3 & 37.3 &24.8 &36.3&48.8 & 11.4 \\
\checkmark & - & \checkmark & 29.5 & 33.9 & 35.1 & \textbf{31.5} &39.9 & 47.1 & 14.7       \\
\checkmark & \checkmark &  \checkmark &\textbf{30.3} &\textbf{34.8} &\textbf{36.1} &\textbf{32.1} &\textbf{40.9} &\textbf{48.3} & \textbf{14.9}\\
\bottomrule
\end{tabular}
}
\end{table}

\begin{table*}[tb]
\centering
\caption{Comparison with state-of-the-art scene graph generation methods on Action Genome~\cite{Action_genome_Ji_2020_CVPR} for PredCls. 
}
\setlength{\tabcolsep}{5pt}
\resizebox{0.7\textwidth}{!}{%
\begin{tabular}{c ccc ccc}
\toprule
  & \multicolumn{3}{c}{With Constraint} & \multicolumn{3}{c}{No Constraints}\\
  \cmidrule(lr){2-4}  \cmidrule(lr){5-7} Method
 & R@10 & R@20 & R@50 & R@10 & R@20 & R@50 \\
\midrule
\midrule
VRD~\cite{lu2016visual}  &51.7  &54.7  &54.7 &59.6 &78.5 &99.2 \\
M-FREQ~\cite{neural_motifs_2017}   &62.4  &65.1 &65.1 &73.4 &92.4 &\textbf{99.6}  \\
MSDN~\cite{Li2017SceneGG} &65.5 &68.5 &68.5 &74.9 &92.7 &99.0 \\
VCTree~\cite{VCTree_Tang_2019_CVPR}  &66.0 &69.3 &69.3  &75.5 &92.9 &99.3 \\
RelDN~\cite{zhang2019vrd} &66.3 &69.5 &69.5 &75.7 &93.0 &99.0 \\
GBS-Net~\cite{lin2020gps} &66.8 & 69.9 &69.9 &76.0 &93.6 &99.5  \\
STTran~\cite{STTran_2021} & 67.3& 70.3& 70.4 & 76.9 &93.9&99.2 \\
{\bf DSG-DETR(Ours)} &\textbf{68.4} &\textbf{71.7} &\textbf{71.7} &\textbf{77.5}&\textbf{94.3}&99.1\\

\bottomrule
\end{tabular}
\label{tab:predcls}}
\end{table*}

\section{Ablation results on SGDet} 
The ablation results on the SGDet task are  summarized in Table \ref{tab:ablation1}.
Similar to the ablation results on SGCls, our relationship transformer shows a performance gain over STTran \cite{STTran_2021} baseline, for the exact same object classification model. 
We observe that the object transformer contributes most (over $4$-points in R@10) to the SGDet performance measured by Recall; while the positional encoding further boosts the performance by $0.8$ in R@10 score under graph-constrained setup. Besides, the improvement in the mean Average Precision (mAP) for object detection also reflects the contribution of each component in DSG-DETR. 
The object transformer exploiting long-term object-consistency improves object detection performance significantly and in turn helps SGDet performance notably.

\section{Results on PredCls}
The edge (predicate) classification conditioned on nodes (subject \& object) is a relatively easy problem to solve. Most recent methods reach the ballpark of 99\% in the unconstrained setup as measured by R@50 for the \textit{PredCls} task as shown in Table \ref{tab:predcls}. The problem becomes challenging when the joint object classification / detection is involved  in SGCls / SGDet tasks which has been the main focus of our main paper.

\section{Long-Tailed Performances}
Although we do not specifically target long-tailed performance in this work, we evaluate the mean Recall for SGCls and SGDet of the proposed DSG-DETR method and compare that with other methods. Mean Recall is computed by averaging recall values over all predicate classes.  
We see in Table ~\ref{tab:mrecall} that DSG-DETR does not compromise on long-tailed performances (as measured by mean Recall values) for achieving state-of-the art SGG performances (as measured by Recall values). 
Another recent work, TRACE ~\cite{TRACE}, utilizes object tracking. However, the object detection backbone of TRACE is based on feature pyramid network, which is known to be superior than the ones used in DSG-DETR and STTran. The efficacy of our method is reflected SGCls performances where all methods use the same backbone.

\begin{table*}[tb]
\centering
\caption{Long-tailed performance evaluation: mean Recall values on Action Genome~\cite{Action_genome_Ji_2020_CVPR} for SGCls and SGDet. 
$\dagger$ denotes evaluating using the official code for the fair evaluation criteria. Please note that SGDet involves object detection. TRACE uses ResNet50-FPN based network for object detection, whereas STTran and DSG-DETR do not use Feature Pyramid Network (FPN) backbone for object detection. 
}
\setlength{\tabcolsep}{2pt}
\resizebox{1\textwidth}{!}{%
\begin{tabular}{c ccccccccc ccccccccc}
\toprule
  & \multicolumn{6}{c}{With Constraint} & \multicolumn{9}{c}{No Constraints}\\
  \cmidrule(lr){2-7} \cmidrule(lr){8-13}
  Method & 
   \multicolumn{3}{c}{SGCls} & \multicolumn{3}{c}{SGDet} &  \multicolumn{3}{c}{SGCls} & \multicolumn{3}{c}{SGDet} \\ \cmidrule(lr){2-4} \cmidrule(lr){5-7} \cmidrule(lr){8-10} \cmidrule(lr){11-13} \cmidrule(lr){14-16}
   & mR@10 & mR@20 & mR@50 & mR@10 & mR@20 & mR@50 & mR@10 & mR@20 & mR@50  & mR@10 & mR@20 &
\\
\midrule
\midrule
TRACE~\cite{zhang2019vrd}$\dagger$ &8.9 &8.9 &8.9 &8.2 &8.5 &8.6 &31.9 &41.2 &43.1 &\bf 22.8 & \bf 30.8 \\
STTran~\cite{STTran_2021} & 27.2& 28.0& 28.0 & \bf 16.6 & \bf 20.8 &\bf 22.2 & 40.7 &50.1 & 58.8 &20.9 & 29.7 \\
{\bf DSG-DETR(Ours)} &\textbf{30.1} &\textbf{30.9} &\textbf{30.9} &16.4 &19.3 &20.1 &\bf 41.0 &\bf 50.9 &\bf 63.2 &21.3 &28.1 \\
\bottomrule
\end{tabular}
\label{tab:mrecall}}
\end{table*}

\section{Tracklet construction mechanism} 

\subsection{Clustering of detections for SGDet}
For SGDet, there could be multiple detections for a single object in one frame, which violates our assumption that each detection is unique for an object in one frame, so we first apply Non-maximum Suppression (NMS) \cite{ren2016faster} to cluster the detections. For each cluster, we select the detection with the highest classification score as the representative. Then we only apply Algorithm
1 (in the main paper)
on the representatives. Once the tracklets obtained, the detections in the same cluster are assigned to the tracklet that their representative belongs to. For the positional encoding, it is first created for a sequence of representatives without considering other detections, according to the sinusoidal encoding \cite{vaswani2017attention}.
\begin{align}
    PE_{(pos, 2i)} &= \sin (pos/10000^{2i/d_{\text{model}}})\\
    PE_{(pos, 2i+1)} &= \cos (pos/10000^{2i/d_{\text{model}}}),
\end{align}
where $pos$ is the position and $i$ is the dimension.  After the creation of the positional encoding for the representatives, the non-representative detections in the tracklet are inserted into the sequence  with the same positional encoding of their representatives. So the detections in one frame share the same positional encoding and $pos$ is invariant to the number of detections in one frame. In SGCls, each detection is a representative since its position is from the ground-truth, and we also use the sinusoidal encoding as the positional encoding.

\subsection{Hyperparameters of coarse tracking algorithm}

The hyperparameters of the coarse tracking algorithm are summarized in Table \ref{tab:hyper}.
\begin{table}[tb!]
\caption{Hyperparameters of Algorithm 
1 (in the main paper)
for SGCls and SGDet. 
}
\label{tab:hyper}
\centering
\begin{tabular}{ccc ccc ccc}
\toprule
Hyperparameter & SGCls & SGDet \\
\midrule
$m$ & 50 & 50&\\
$\lambda_{\text{feat}}$ & 2 & 0\\
$\lambda_{\text{iou}}$ & 1 & 0\\
$\lambda_{L_1}$ & 2 & 0 \\
$\tau$ & 0.5 & 0.5\\
\bottomrule
\end{tabular}
\end{table}

Note for SGDet, we find that simply placing the detections with the same predicted labels from Faster R-CNN \cite{ren2016faster} in a sequence could achieve a good enough result with fast computation. Although the results may not be tuned to the best, they basically augment two arguments we proposed previously: (1) DSG-DETR can still take effect with very coarse tracking sequences, (2) the object transformer of DSG-DETR reasons over long-term temporal dependencies beyond a majority voting.

\section{Details of model architecture}
\subsection{Object transformer}
The object transformer consists of three transformer encoder layers as in \cite{vaswani2017attention}. For its input, the embedding functions $g^{\text{box}}$ and $g^{\text{dist}}$ are defined  as
\begin{align}
    g^{\text{box}} &= \text{Dropout}(\text{ReLU}(\mathbf{W}^{\text{box}}\text{BatchNorm}(\mathbf{b}^T)))\\
    g^{\text{dist}} &= \mathbf{W}^{\text{dist}}\tilde{\mathbf{c}}^T,
\end{align}
where $\mathbf{W}^{\text{box}}\in\mathbb{R}^{4\times 128}$ and $\mathbf{W}^{\text{dist}}\in\mathbb{R}^{35\times 200}$, and we set $\text{Dropout}$ with rate $0.1$. The self-attention module of the object transformer has $8$ heads with $d_{\text{model}}=2376$ and $\text{Dropout}$ rate $0.1$, its output is then passed to a feedforward (FFN) network with hidden size $2048$. The output of the transformer encoder is further passed through a two-layer FFN network with hidden size $1024$ and output dimension the number of classes.

\subsection{Relationship transformer}
The spatial transformer encoder consists of one transformer encoder layer and the temporal transformer encoder consists of three transformer encoder layers. Similar to STTran \cite{STTran_2021}, we use the following embedding functions for the relationship representations,
\begin{align}
    g^s(\mathbf{f}_s) &= \mathbf{W}^s\mathbf{f}_s\\
    g^o(\mathbf{f}_o) &= \mathbf{W}^o\mathbf{f}_o\\
    g^{sp}(\mathbf{u}_{so}\oplus g^{\text{boxes}}(\mathbf{b}_s, \mathbf{b}_o)) &= \mathbf{W}^{sp}\phi(\mathbf{u}_{so}\oplus g^{\text{boxes}}(\mathbf{b}_s, \mathbf{b}_o))\\
    g^{se}(\mathbf{c}) &= \mathbf{W}^{glove}\mathbf{c}^T.
\end{align}
Here, $\mathbf{W}^s, \mathbf{W}^o \in \mathbb{R}^{2048\times 512}$, $\mathbf{W}^{sp}\in\mathbb{R}^{12544\times 512}$ and $\mathbf{W}^{glove}\in \mathbb{R}^{35\times 200}$, which is the object class embedding initialized with GloVe word embeddings \cite{pennington-etal-2014-glove}. $\mathbf{u}_{so}\in\mathbf{R}^{256\times7\times7}$ is the union feature of the subject $s$ and object $o$, extracted by RoIAlign \cite{he2018mask}, $g^{box}$ is a function projects the bounding boxes of the subject and object to the same dimension as $\mathbf{u}_{so}$ and $\phi$ is the flattening operation.

The spatial encoder and temporal encoder layers share the similar architecture as the object transformer encoder layer except that $d_{\text{model}}=1936$. The linear projections on the predicate logits have the projection matrix shape $\mathbf{W}^{atten}\in\mathbb{R}^{1936\times 3}$,  $\mathbf{W}^{spatial}\in\mathbb{R}^{1936\times 6}$ and  $\mathbf{W}^{contact}\in\mathbb{R}^{1936\times 16}$, with the Softmax head applied on the attention logits and Sigmoid head on the spatial and contact logits.

\end{document}